\documentclass{tlp}

\usepackage{amsmath}
\usepackage{amssymb}

\usepackage[utf8]{inputenc}
\usepackage{microtype}

\usepackage{url}
\usepackage{stmaryrd}
\usepackage{listings}

\providecommand{\sysfont}{\textit}

\newcommand{\Clingo}{\sysfont{Clingo}}

\newcommand{\Clingcon}{\sysfont{Clingcon}}
\newcommand{\clingcon}{\sysfont{clingcon}}
\newcommand{\clingo}{\sysfont{clingo}}

\newcommand{\flingo}{\sysfont{flingo}}
\newcommand{\Flingo}{\sysfont{Flingo}}

\newcommand{\lctocasp}{\sysfont{lc2casp}}

\providecommand{\C}{C}

\providecommand{\logfont}{\textrm}

\newcommand{\HTC}{\ensuremath{\logfont{HT}_{\!c}}}

\providecommand{\Underscore}{\textunderscore}

\lstdefinelanguage{clingo}{%
  basicstyle=\ttfamily,%
  keywordstyle=[1]\bfseries,%
  keywordstyle=[2]\bfseries,%
  keywordstyle=[3]\bfseries,%
  showstringspaces=false,%
  literate={_}{\Underscore}1 {\%\%}{}0,%
  escapeinside={\#(}{\#)},%
  alsoletter={\#,\&},%
  keywords=[1]{not,from,import,def,if,else,elif,return,while,break,and,or,for,in,del,and,class,with,as,is,yield,async},%
  keywords=[2]{\#const,\#show,\#minimize,\#base,\#theory,\#count,\#external,\#program,\#script,\#end,\#heuristic,\#edge,\#project,\#show,\#sum},%
  morecomment=[l]{\#\ },%
  morecomment=[l]{\%\ },%
  morestring=[b]",%
  stringstyle={\itshape},%
  commentstyle={\color{darkgray}}%
}

\lstdefinelanguage{clingcon}[]{clingo}{morekeywords={&dom,&sum,&nsum,&diff,&disjoint,&distinct,&minimize,&maximize,&show}}
\lstdefinelanguage{flingo}[]{clingo}{morekeywords={&sum,&sus,&in,&df,&min,&max,&show}}
\lstdefinelanguage{clingodl}[]{clingo}{morekeywords={&diff}}

\lstdefinelanguage{python}{%
  basicstyle=\ttfamily,%
  keywordstyle=[1]\bfseries,%
  showstringspaces=false,%
  literate={_}{\Underscore}{1},%
  escapeinside={\#(}{\#)},%
  alsoletter={\#,\&},%
  keywords=[1]{not,from,import,def,if,else,elif,return,while,break,and,or,for,in,del,and,class,with,as,is,yield,async},%
  morecomment=[l]{\#\ },%
  morestring=[b]",%
  stringstyle={\itshape},%
  commentstyle={\color{darkgray}}%
}

\renewcommand{\sysfont}[1]{\texttt{#1}}
\newcommand{\clingof}{\clingo\{\sysfont{f}\}}

\lstdefinelanguage{clingos}{%
  language=clingo,%
  basicstyle=\small\ttfamily%
}

\newtheorem{proposition}{Proposition}

\newtheorem{definition}{Definition}

\newcommand{\den}[1]{\llbracket \, #1 \, \rrbracket}
\newcommand{\ctermm}[3]{\ensuremath{{#1|#2}{:\,}#3}}
\newcommand{\cterm}[3]{\ensuremath{(\ctermm{#1}{#2}{#3})}} 

\newcommand{\eval}[2]{\ensuremath{\mathit{eval}}_{\langle #1,#2\rangle}}

\newcommand{\eqdef}{%
  \mathrel{\vbox{\offinterlineskip\ialign{%
    \hfil##\hfil\cr%
    $\scriptscriptstyle\mathrm{def}$\cr%
    \noalign{\kern1pt}%
    $=$\cr%
    \noalign{\kern-0.1pt}%
}}}}

\newcommand{\code}[1]{\ensuremath{\texttt{#1}}}

\newcommand{\tuple}[1]{\langle #1 \rangle}

\newcommand{\X}{\ensuremath{\mathcal{X}}}
\newcommand{\D}{\ensuremath{\mathcal{D}}}

\newcommand{\T}{\ensuremath{\mathcal{T}}}

\renewcommand{\C}{\ensuremath{\mathcal{C}}}

\newcommand{\V}{\ensuremath{\mathcal{V}}}

\newcommand{\vars}[1]{\ensuremath{\mathit{vars}(#1)}}

\newcommand{\true}{\ensuremath{\mathbf{t}}} 
\newcommand{\undefined}{\ensuremath{\boldsymbol{u}}} 

\newcommand{\Z}{\ensuremath{\mathbb{Z}}}

\newcommand{\p}{\ensuremath{\mathtt{p}}}

\def\susf{\ensuremath{\mathtt{sus}}}
\def\sumf{\ensuremath{\mathtt{sum}}}
\def\maxf{\ensuremath{\mathtt{max}}}
\def\minf{\ensuremath{\mathtt{min}}}

\newcommand{\Susf}{\normalfont{\texttt{\&sus}}}
\newcommand{\Sumf}{\normalfont{\texttt{\&sum}}}
\newcommand{\Maxf}{\normalfont{\texttt{\&max}}}
\newcommand{\Minf}{\normalfont{\texttt{\&min}}}
\newcommand{\Deff}{\normalfont{\texttt{\&df}}}

\newcommand{\defh}{\normalfont\texttt{\&df}}
\newcommand{\inth}{\normalfont\texttt{\&int}}

\newcommand{\Sum}[1]{\code{\&sum\{} #1 \code{\}}}
\def\Leq{\mbox{\tt <=}}
\def\Geq{\mbox{\tt >=}}
\def\Neq{\mbox{\tt !=}}
\def\Grt{\mbox{\tt >}}
\def\Low{\mbox{\tt <}}
\def\Equ{\mbox{\tt =}}
\newcommand{\cl}{\mathit{cl}}
\newcommand{\fl}{\mathit{fl}}
\newcommand{\Not}{\mathtt{not}}
\newcommand{\smin}{\ensuremath{\mathit{min}}}
\newcommand{\smax}{\ensuremath{\mathit{max}}}
\newcommand{\Zcl}{\ensuremath{\mathbb{Z}_\cl}}
\newcommand{\Dcl}{\ensuremath{\D_{\!\cl}}} 
\newcommand{\Vcl}{\ensuremath{\V_{\!\cl}}} 
\newcommand{\Ass}{\code{=:}}
\newcommand{\ruleo}{\code{:-}}

\lstdefinestyle{ASP}{
    basicstyle=\ttfamily\footnotesize,
    breakatwhitespace=false,
    breaklines=true,
    captionpos=b,
    keepspaces=true,
    numbers=left,
    numbersep=5pt,
    showspaces=false,
    showstringspaces=false,
    showtabs=false,
    tabsize=2
}

\lefttitle{Pedro Cabalar et al.}

\jnlPage{1}{8}
\jnlDoiYr{2026}
\doival{10.1017/xxxxx}

\begin{document}
\title{\flingo\ --- Instilling ASP Expressiveness into Linear Integer Constraints}
\begin{authgrp}
\author{\sn{Pedro} \gn{Cabalar}}
\affiliation{University of Corunna, Spain}
\author{\sn{Jorge} \gn{Fandinno}}
\affiliation{University of Nebraska Omaha, USA}
\author{\sn{Torsten} \gn{Schaub}}
\affiliation{University of Potsdam, Germany}\affiliation{Potassco Solutions, Germany}
\author{\sn{Philipp} \gn{Wanko}}
\affiliation{Potassco Solutions, Germany}
\end{authgrp}
\maketitle
%
\begin{abstract}
Constraint Answer Set Programming (CASP) is a hybrid paradigm that enriches Answer Set Programming (ASP) with numerical constraint processing,
a crucial requirement for many real-world applications.
However, the specification of constraints in most CASP solvers aligns more closely with the expressiveness and semantics of the numerical back-end than the ASP paradigm.
In the latter, numerical attributes are represented with predicates and this allows for declaring default values, leaving the attribute undefined, making non-deterministic assignments with choice rules or using aggregated values.
In CASP, most (if not all) of these features are lost once we switch to a constraint-based representation of those same attributes.
In this paper, we present the \flingo\ language (and tool) that incorporates the aforementioned expressiveness inside the numerical constraints and we illustrate its use with several examples.
Based on previous work that established its semantic foundations, we also present a translation from the newly introduced \flingo\ syntax to regular CASP programs following the \clingcon\ input format.
\end{abstract}
%

%
%
\section{Introduction}\label{sec:introduction}

One of the extensions of \emph{Answer Set Programming} (ASP;~\citealt{lifschitz99a})
that is frequently demanded for practical applications is \emph{Constraint ASP} (CASP;~\citealt{megezh08a,lierler23a}).
Many domains require efficient numerical computations that exceed the capabilities of state of the art ASP tools for combinatorial problem solving.
These tools typically divide their computation into a \emph{grounding} phase (removing variables by instantiation) and a \emph{solving} phase (getting the answer sets of a propositional program).
In this schema, numerical constraints are solved by considering all the possible instantiations of the involved numerical variables during the grounding phase.
This strategy has a serious problem of scalability: a simple change of measurement in some quantity (e.g., moving from
meters to centimeters) causes a blow-up in the grounding of the program that immediately leads to intractability.
The basic idea of CASP is delegating numerical computation to a back-end constraint solver,
following the spirit of \emph{Satisfiability Modulo Theory} (SMT;~\citealt{niolti05a}).
Thus, the ASP program is extended with \emph{constraint atoms} that express the numerical restrictions associated with the problem at hand.

While this method mitigates grounding inefficiencies,
it introduces a significant modeling challenge:
the specification of constraint atoms often aligns more closely with the operational requirements of the numerical backend
than with the standard ASP paradigm.
Consequently, several high-level declarative features,
such as default value declarations, undefined attributes, non-deterministic assignments via choice rules, and complex aggregates,
are no longer directly accessible when these attributes are represented as constraint atoms.
Just to put a simple example, consider the recent situation in which the default tariff for products from EU in the USA is fixed to 15\%.
This is a default that may have exceptions, such as the case of steel (with a tariff of 25\%) or aircraft products, that are exempt.
A possible representation%
\footnote{The \clingo\ conditional literal ``\code{\#false: tariff(P,eu,T), T!=15}'' can be replaced by a negated
  auxiliary predicate ``\code{not other\_eu\_tariff(P)}'' defined with rule ``\code{other\_eu\_tariff(P) :- tariff(P,eu,T), T!=15}''.}
of this problem in \clingo\ could look like:
\begin{lstlisting}[language=clingos,numbers=none]
tariff(P,eu,15) :- sales(P,eu,_), #false: tariff(P,eu,T), T!=15.
tariff(steel,eu,0).
tariff(aircraft,eu,25).
\end{lstlisting}
where predicate \code{sales(P,C,X)} reflects the total amount \code{X} of sales of products from country (or market) \code{C} of type \code{P}.
Moreover, suppose we want to collect the total income from taxes as follows:
\begin{lstlisting}[language=clingos,numbers=none]
taxincome(Z) :- #sum{X*T/100,P,C:sales(P,C,X),tariff(P,C,T)}=Z.
\end{lstlisting}
\noindent If we decide to represent the tariff as a numerical constraint variable, default values are usually lost, since reasoning with constraint atoms is typically monotonic.
Additionally, we cannot compute the tax income with an ASP aggregate, since the latter is defined in terms of logical variables which, in general, cannot retrieve computations from numerical constraints (otherwise, the grounding gain is lost).
Finally, the ASP code above can easily accommodate lack of information: for instance, our database may contain a fact like {\tt sales(cars,us,30000)} without having to define any tariff for the USA itself, or may include information about the tariff for Canadian cars {\tt tariff(cars,ca,25)} without knowing the tariff for the rest of products from that country.
Declaring tariffs as constraint variables {\tt tariff(steel,ca)}, {\tt tariff(food,ca)}, etc  forces us to explicitly assign them numerical values.

One approach in the literature that allowed undefinedness and default values for constraint variables was the logic of \emph{Here-and-There with constraints} (\HTC) proposed by \cite{cakaossc16a}.
In that work, a first prototype \lctocasp\ was presented as a proof of concept, but was still far from extrapolating standard ASP representational features to constraints.
Later on, another of those common ASP features, aggregates, was also incorporated into constraint atoms~\citep{cafascwa20a, cafascwa20b} under a semantics based on \HTC, although a proper language definition and the corresponding tool implementation was still missing.

In this paper,
we introduce the \flingo\ system,
a comprehensive extension of \clingo\ that enables the specification of
undefined variables, default values, non-deterministic choices, and aggregates within constraint atoms.
These features are handled uniformly with respect to the standard ASP paradigm,
maintaining a consistent modeling experience.
We formalize the semantics of the \flingo\ language in terms of \HTC,
illustrate its application through various examples, and
provide a translation into regular CASP programs following the \clingcon\ input format~\citep{bakaossc16a}.

The rest of the paper is organized as follows.
The next section contains the background, including first an overview of \HTC, followed by a brief description of systems \clingo\ and \clingcon, on which \flingo\ is based.
Section~\ref{sec:approach} describes the \flingo\ language, defining its input syntax, the associated semantics and some useful abbreviations.
In Section~\ref{sec:flingo:system},
we provide details about the implementation of the \flingo\ solver, which mainly relies on a translation into \clingcon, used as a back-end.
Finally, Section~\ref{sec:discussion} summarizes our paper.


\section{Background}\label{sec:background}
\subsection{The Logic of Here-And-There with Constraints}
The logic of \emph{Here-and-there with constraints} (\HTC;~\citealt{cakaossc16a}) is an extension of the logic of Here-and-There (HT;~\citealt{heyting30a}).
Analogously to Equilibrium Logic~\citep{pearce96a},
a model selection is defined on \HTC\ models,
which provides the logical foundations for constraint satisfaction problems (CSPs) in the setting of ASP.
In \HTC, a CSP is expressed as a triple $\tuple{\X,\D,\C}$, also called \emph{signature},
where \X\ is a set of \emph{variables} over some non-empty \emph{domain} \D,
and \C\ is a set of \emph{constraint atoms}.
%
%
We assume a partition of the set of variables~$\X$ into two disjoint sets:
\emph{propositional},~$\X^p$, and~\emph{non\nobreakdash-propositional variables},~$\X^n$.
Similarly, the set of constraint atoms~$\C$ is also partitioned into two disjoint sets:
\emph{propositional},~$\C^p$, and \emph{non\nobreakdash-propositional atoms},~$\C^n$.
The set of propositional atoms~$\C^p$ contains a constraint atom~$\p$ for each propositional variable~$p\in\X^p$,
that is,~$\C ^p = \{ \p \mid p \in \X^p\}$.
Note that we use different fonts to distinguish the propositional variable~$p$ from the propositional atom~$\p$.
We also assume that the domain~$\D$ contains the element~$\true$, which is used to denote the truth value \emph{true} of propositional variables.
A \emph{propositional formula} over~$\C^p$ is defined by the grammar:
\begin{gather}
    \varphi ::= \bot \ \mid \ \p\in \C^p \ \mid \ \varphi \land \varphi \ \mid \ \varphi \lor \varphi \ \mid \ \varphi \to \varphi \ .
\end{gather}

Non-propositional atoms can be arbitrary strings, in which we distinguish substrings called \emph{terms}.
Terms are divided into \emph{basic} and \emph{conditional terms}.
We assume a set of basic terms~$\T^b$ that contains all the variables, domain elements and the special symbol~$\undefined$, representing an undefined value.
A \emph{conditional term} is an expression of the form
\begin{gather}
    \cterm{s_1}{s_2}{\varphi}
\end{gather}
where~$s_1$ and~$s_2$ are basic terms, and~$\varphi$ is a propositional formula.
As an example,
\begin{gather}
    \sumf\{ \ \cterm{3*x}{1}{p} \ ; \ \cterm{2}{x}{q} \ \} < 3
\end{gather}
is a non-propositional atom, where~$x$ is a variable,~$1$ and~$2$ are domain elements, and~$p$ and~$q$ are propositional variables.
The variables, the domain elements and the string~$3*x$ are basic terms.
$\cterm{3*x}{1}{p}$ and $\cterm{2}{x}{q}$ are conditional terms.
Intuitively, the conditional term~$\cterm{3*x}{1}{p}$ means that if~$p$ is true, then the value of the term is~$3*x$, otherwise it is~$1$.
Similarly, the conditional term~$\cterm{2}{x}{q}$ means that if~$q$ is true, then the value of the term is~$2$, otherwise its value is~$x$.
The value of the two conditional terms is then added up (\sumf) and compared to~$3$.

A \emph{formula} over~$\C$ is defined by the grammar:
\begin{gather}
    \varphi ::= \bot \ \mid \ c \in \C \ \mid \ \varphi \land \varphi \ \mid \ \varphi \lor \varphi \ \mid \ \varphi \to \varphi \ .
\end{gather}
Hence, all propositional formulas are also formulas.
As usual, we define
$\top$ as~$\bot\to\bot$,
$\neg\varphi$ as~$\varphi\to\bot$, and
let~$\varphi_1\leftrightarrow\varphi_2$ stand for~${(\varphi_1\to\varphi_2) \land (\varphi_2\to\varphi_1)}$.
For every expression~$e$, we let~$\vars{e}$ denote the set of non-propositional variables occurring in~$e$.

For the semantics,
we start by defining an extended domain as ${\mathcal{D}_{\undefined} \eqdef \mathcal{D} \cup \{\undefined\}}$.
A \emph{valuation} $v$ over $\mathcal{X},\mathcal{D}$ is a function
\mbox{$v:\mathcal{X}\rightarrow\mathcal{D}_{\undefined}$}
where
\mbox{$v(x)=\undefined$}
represents that variable~$x$ is left undefined.
Moreover, if ${X \subseteq \mathcal{X}}$ is a set of variables, valuation
${v|_X: X\rightarrow\mathcal{D}_{\undefined}}$ stands for the projection of $v$ on~$X$.
A valuation $v$ can alternatively be seen as the set of pairs
\mbox{$\{ (x \mapsto v(x)) \mid x \in \mathcal{X}, v(x)\in\mathcal{D}\}$}
so that no pair in the set maps a variable to~$\undefined$.
This representation allows us to use standard set inclusion for comparison.
We thus write ${v\subseteq v'}$ to mean that~$v$ and~$v'$ agree on all variables defined by~$v$, but~$v'$ may define additional variables.
By~$\mathcal{V}$ we denote the set of valuations over~$\X,\D$.

A constraint atom or formula is \emph{basic} if it does not contain conditional terms.
By~$\C^b$ we denote the set of basic constraint atoms.
The semantics of basic constraint atoms is defined by a denotation~$\den{\cdot} : \C^b \to 2^{\mathcal{V}}$ mapping each basic constraint atom~$c$ to the set of valuations satisfying~$c$.
For each propositional atom~$\p \in \C^p$, we assume that its denotation satisfies
\(
\den{\p} \ \eqdef \ \{ v \in \mathcal{V} \mid v(p) = \true \}
\).
In \HTC,
an \emph{ht-interpretation} over $\X,\D$ is a pair $\langle h,t \rangle$
of valuations over $\X,\D$ such that $h\subseteq t$.
%
\begin{definition}\label{def:basicsat}
  Given a denotation $\den{\cdot}$,
  an interpretation $\langle h,t \rangle$ \emph{satisfies} a formula~$\varphi$,
  written $\langle h,t \rangle \models \varphi$,
  if
  \begin{enumerate}
    \item $\langle h,t \rangle \models c \text{ if }  h\in \den{\eval{h}{t}(c)}$    \label{item:htc:atom}
    \item $\langle h,t\rangle \models \varphi \land \psi \text{ if }  \langle h,t\rangle \models \varphi \text{ and }  \langle h,t\rangle \models \psi$
    \item $\langle h,t\rangle \models \varphi \lor \psi \text{ if }  \langle h,t\rangle \models \varphi \text{ or }  \langle h,t\rangle \models \psi$
    \item $\langle h,t\rangle \models \varphi \rightarrow \psi
    \text{ if }\langle w,t\rangle \not\models \varphi \text{ or } \langle w,t\rangle \models \psi
    \text{ for all }w\in\{h,t\}$
  \end{enumerate}
  where~$\eval{h}{t}(c)$ is the result of
  replacing each conditional term of the form~$\cterm{s_1}{s_2}{\varphi}$ in~$c$ with~$\eval{h}{t}\cterm{s_1}{s_2}{\varphi}$,
  which is defined as follows:
  \begin{align*}
    \eval{h}{t}\cterm{s_1}{s_2}{\varphi}
    \ &\eqdef \
      \left\{
      \begin{array}{ll}
        s_1 & \text{if } \langle h,t\rangle\models\varphi\\
        s_2 & \text{if } \,\langle t,t\rangle \not\models\varphi\\
        \undefined & \text{otherwise}
      \end{array}
      \right.
  \end{align*}
\end{definition}
%
When~$\langle h,t \rangle \models \varphi$ holds, we say that~$\langle h,t \rangle$ is an \emph{ht-model} of~$\varphi$.
We write~$t \models \varphi$ if~$\langle t,t \rangle \models \varphi$ and we say that~$t$ is a \emph{model} of~$\varphi$.
A (ht-)model of a set~$\Gamma$ of formulas is a (ht-)model of all formulas in~$\Gamma$.
%
A model~$t$ of a set~$\Gamma$ is a \emph{stable model} of~$\Gamma$ if there is no~$h \subset t$ such that~$\langle h,t \rangle \models \Gamma$.
%

\subsection{ASP System \clingo\ and CASP System~\clingcon}
\Clingo~\citep{gekakaosscwa16a} is an ASP system that provides a powerful interface for integrating external theories.
It allows us to extend the basic ASP solver with new theory atoms by providing a theory grammar defining their syntax,
and an implementation of their semantics via so-called \emph{theory propagators}.
The CASP system~\clingcon~\citep{bakaossc16a} uses this \clingo\ interface to enrich the ASP language with
linear constraint atoms over integers of the form
\begin{gather}
    \Sum{k_1 * x_1 ; \ldots ; k_n * x_n } \prec k_0
    \label{eq:clingcon-atom}
\end{gather}
where each~$k_i$ is an integer, each~$x_i$ is a (non-propositional) integer variable, and $\prec$ is one of the relations~\Leq, \Equ, \Neq, \Low, \Grt, \Geq.
Recently, it has been shown that the semantics of many of the systems built using \clingo's theory interface,
including \clingcon,
can be nicely described in terms of~\HTC\ by associating a denotation with each theory atom~\citep{cafascwa25a}.
\Clingcon\ assumes a finite%
\footnote{The current implementation assumes the domain of integers between~$\min_\cl$ and~$\max_\cl$
  that can be set using command line arguments~$\texttt{--min--int=<i>}$ and~$\texttt{--max-int=<i>}$.
  The default values are~$-2^{30}+1$ and~$2^{30}-1$, which are also the minimum and maximum values that can be selected.}
domain of integers~$\Zcl \subseteq \mathbb{Z}$ with a minimum,~$\smin_\cl$, and a maximum value,~$\smax_\cl$.
Consequently, the \HTC\ domain to capture \clingcon's semantics is~$\Dcl  = \Zcl \cup \{\true\}$.
The set of all valuations with domain~$\Dcl$ is given by~$\Vcl$.
With it, the denotation of atom~\eqref{eq:clingcon-atom} is defined as follows:
\begin{gather}
    \textstyle\big\{
    v\in\Vcl \mid
    v(x_i) \in \mathbb{Z}\text{ for } 1\leq i \leq n
    \text{ and }
    \sum_{1\leq i \leq n}k_i*v(x_i)\prec k_0
    \big\}
    \label{eq:clingcon-denotation}
\end{gather}

Let us now recall the syntax of \clingcon\ programs~\citep{bakaossc16a}.
A \clingcon~\emph{atom} is either a propositional atom or a constraint atom of form~\eqref{eq:clingcon-atom}.
A \clingcon~\emph{literal} is any \clingcon\ atom~$a$ or its single or double negation, $\Not\ a$ or $\Not\ \Not\ a$, respectively.
A \clingcon~\emph{rule} is an expression of the form
\begin{gather}
    a \ \code{:-} \ l_1, \ldots, l_n
    \label{eq:clingcon-rule}
\end{gather}
where~$a$ is a \clingcon\ atom or the symbol~$\bot$ denoting \emph{falsity}, and each~$l_i$ is a \clingcon\ literal.
An expression of the form
\begin{gather}
    \{ \code{p} \} \ \code{:-} \ l_1, \ldots, l_n
    \label{eq:clingcon-choice-rule}
\end{gather}
where~$\code{p}$ is a propositional atom, is called a \emph{choice rule}, and it is an abbreviation for the rule
`\(
\code{p} \, \code{:-} \, l_1, \ldots, l_n,\, \Not\,\Not\ \code{p}
\)'.
%
Moreover,
\clingcon\ accepts all the constructs of \clingo, including disjunctive rules, aggregates, and weak constraints.

We define the semantics of \clingcon\ programs via a translation $\tau$ into \HTC.
For each \clingcon\ literal~$l$, we denote by~$\tau l$ the corresponding \HTC\ literal, that is,
$\tau (\Not\ \Not\ a) \eqdef \neg \neg a$, $\tau (\Not\ a) = \neg a$ and $\tau a \eqdef a$ for any atom $a$.
For each rule~$r$ of form~\eqref{eq:clingcon-rule}, we denote by~$\tau r$ the formula
\begin{gather}
    \tau l_1 \land \ldots \land \tau l_n \to \tau a
    \label{eq:tau.clingcon-rule}
\end{gather}
Furthermore, we need to introduce new constraints in~$\HTC$, which are not available in \clingcon.
These constraint atoms are of the form~$\defh(x)$ and~$\inth(x)$, where~$x$ is a variable.
These atoms are associated with the following denotations:
\begin{align*}
    \den{\defh(x)} &\ \eqdef \ \{ v \in \Vcl \mid v(x)\neq\undefined \} \\ 
    \den{\inth(x)} &\ \eqdef \ \{ v \in \Vcl \mid v(x) \in \Z_\cl \}
\end{align*}
%
%
For a set~$P$ of \clingcon\ rules, we define~$\cl(P)$ as the set of \HTC\ formulas containing formula~$\tau r$ for each rule~$r \in P$
as well as formulas of the form
\begin{align}
    &\defh(p) \to \p &&\quad \text{for each propositional variable } p \in \X^p
    \label{eq:clingcon-defh}
    \\
    &\inth(x)        &&\quad \text{for each integer variable } x \in \X^n
    \label{eq:clingcon-inth}
\end{align}
Formulas~\eqref{eq:clingcon-defh} and~\eqref{eq:clingcon-inth} ensure that variables only take their intended values:
propositional variables are either undefined or take the value~$\true$, while integer variables are always defined and can take any integer value.
There is a one-to-one correspondence between the stable models of a \clingcon\ program~$P$ and the stable models of the \HTC\ theory~$\cl(P)$
as shown in~\citep[Theorem~4]{cafascwa25a}.
%
%

\section{The \flingo\ language}\label{sec:approach}

\Flingo\ is a CASP system that, much like \clingcon, enriches the ASP language with integer constraint atoms.
The main feature of \flingo\ is that the values assigned to integer variables need to be \emph{founded}.
Recall that in ASP,
propositional atoms are \emph{false by default} and
true propositional atoms need to be founded~\citep{gerosc91a}, that is, derived from the facts via the rules of the program.
Analogously, in \flingo, integer variables are \emph{undefined by default} and any defined integer variable needs to
occur in a founded constraint atom.
This behavior is inspired by \HTC\ and is significantly different from other existing CASP solvers.
As an example, consider the program\footnote{Both \clingcon\ and \flingo\ (currently) rely on \clingo's theory grammar,
  which makes us express $x=1$ as $\code{\&sum\{$x$\} = 1}$.}
\begin{align}
    \code{\{a\}.}   &&&\code{\&sum\{$x$\} = 1 :- a.}
    \label{eq:flingo-example}
\end{align}
which has exactly two stable models in \flingo:
one where~$\code{a}$ is false and~$x$ is undefined, and another where~$\code{a}$ is true and~$x$ is assigned the value~$1$.
Both of these are also \clingcon\ stable models,
but \clingcon\ has infinitely many stable models where~$\code{a}$ is false and~$x$ is assigned any integer.
The second distinct feature of \flingo\ is its support of \emph{aggregates} over integer variables,
something that is not supported by other existing CASP solvers.

\subsection{Syntax}
As mentioned above, \flingo\ supports an enriched syntax compared to \clingcon.
We start by describing \flingo's \emph{conditional terms},
which are inspired by aggregate elements in ASP and conditional terms in \HTC.
A \emph{product term} is either an integer, an integer variable, or an expression of the form~$n * x$ where~$n$ is an integer and~$x$ is an integer variable.
In \flingo, a \emph{conditional term} is an expression of the form
\begin{gather}
    s : l_1, \ldots, l_n
    \label{eq:conditional.term}
\end{gather}
where~$s$ is a product term and~$l_1, \ldots, l_n$ is a comma separated list of propositional literals.
There are two main differences with respect to conditional terms in \HTC.
First, the condition is restricted to be a conjunction of propositional literals instead of an arbitrary propositional formula,
as common in ASP.
Second, conditional terms have no alternative term in \flingo.
The term~$s$ is the one considered when the condition~$l_1, \ldots, l_n$ is satisfied, otherwise a default term is used.
This default term is context-dependent and discussed below with the semantics of \flingo.
With these changes, we bring \HTC\ conditional terms closer to the usual logic programming syntax for aggregates.

A \flingo\ \emph{term} is either a product term or a \flingo\ conditional term.
%
The language of~\flingo\ supports the following constraint atoms:
\begin{align}
    &\Sumf\{t_1 ; \ldots ; t_n\} \prec s
    \label{eq:flingo-sumf}
    \\
    &\Susf\{t_1 ; \ldots ; t_n\} \prec s
    \label{eq:flingo-susf}
    \\
    &\Minf\{t_1 ; \ldots ; t_n\} \prec s
    \label{eq:flingo-maxf}
    \\
    &\Deff\{x\}
    \label{eq:flingo-deff}
\end{align}
where each~$t_i$ is a \flingo\ term, $s$ is a product term, $x$ is an integer variable, and $\prec$ is one of the following relations~\Leq, \Equ, \Neq, \Low, \Grt, \Geq.
The operation~$\susf$ stands for~\emph{strict sum}
(more precisely, $\susf$ stands for \texttt{su}\emph{m} \texttt{s}\emph{trict}),
while~$\sumf$ is its non-strict version, similar to the one used in ASP.
Intuitively, the non-strict version of sum discards undefined variables, while the strict does not.
If any variable is undefined, the constraint atom with strict sum is false.
As an example, consider the valuation~${v = \{ x_1 \mapsto 1 \}}$ and the constraint atoms
\begin{align}
    \Sumf\{x_1 ; x_2\} \ \Leq \ 3
    \label{eq:flingo-sumf-example}
    \\
    \Susf\{x_1 ; x_2\} \ \Leq \ 3
    \label{eq:flingo-susf-example}
\end{align}
In this valuation, variable~$x_2$ is undefined.
Atom~\eqref{eq:flingo-sumf-example} is satisfied by valuation~$v$ because~$v(x_1) = 1 \leq 3$,
while~$x_2$ is undefined
and thus discarded from the sum.
In contrast, atom~\eqref{eq:flingo-susf-example} is not satisfied simply because~$x_2$ is undefined.
Operation~$\minf$ always work in a non-strict way:
for instance, $\minf$ always returns the minimum value among the defined terms in the set.

A \flingo\ \emph{atom} is either a propositional atom or a constraint atom of the forms~(\ref{eq:flingo-sumf}--\ref{eq:flingo-deff}).
A \flingo\ \emph{literal} is any \flingo\ atom~$a$ or its single or double default negation, $\Not\ a$ or $\Not\ \Not\ a$, respectively.
A \flingo\ \emph{rule} is an expression of the form
\begin{gather}
    a \ \code{:-} \ l_1, \ldots, l_n
    \label{eq:flingo-rule}
\end{gather}
where~$a$ is a \flingo\ atom or the symbol~$\bot$ denoting \emph{falsity}, and each~$l_i$ is a \flingo\ literal.
Note that, syntactically, every \clingcon\ rule is also a \flingo\ rule.

\subsection{Semantics}
The semantics of \flingo\ is defined using~\HTC\ similar to \clingcon\
but it uses a different denotation for constraint atoms, and a slightly different translation.

We start by defining a \emph{neutral element} for each constraint atom operation
\begin{align*}
    0^\sumf &\eqdef 0 &\qquad
    0^\susf &\eqdef 0 &\qquad
    0^\maxf &\eqdef \smin_\cl &\qquad
    0^\minf &\eqdef \smax_\cl
\end{align*}
Both sum operations use~$0$ as their neutral element,
while~$\maxf$ and~$\minf$ use the minimum and maximum elements of the integer domain, respectively.
We define the denotation of the basic constraint atoms as follows:
\begin{align*}
    \den{\Sumf\{t_1 ; \ldots ; t_n \} \prec s} &\eqdef
    \big\{ v \in \Vcl \mid
           v^{\sumf}(t_i) \in \Z\text{ for } 1\leq i \leq n
           , 
           \textstyle\sum_{1\leq i \leq n} v^{\sumf}(t_i) \prec v(s)
    \big\}
    \\
    \den{\Susf\{t_1 ; \ldots ; t_n \} \prec s} &\eqdef
    \big\{ v \in \Vcl \mid
           v(t_i) \in \Z\text{ for } 1\leq i \leq n
           , 
           \textstyle\sum_{1\leq i \leq n} v(t_i) \prec v(s)
    \big\}
    \\
    \den{\Minf\{t_1 ; \ldots ; t_n \} \prec s} &\eqdef
    \big\{ v \in \Vcl \mid
        \textstyle\min_{1 \leq i \leq n} v^{\minf}(t_i) \prec v(s)  \big\}
    \\
    \den{\Deff(x)} &\eqdef
    \big\{ v \in \Vcl \mid v(x)\neq\undefined \big\}
\end{align*}
where
\begin{align*}
    v(n)     &\eqdef n
    &\hspace{5pt}
    v(n * x) &\eqdef
        \begin{cases} n * v(x)   &\text{if } v(x) \in \Z \\
                      \undefined &\text{otherwise}
        \end{cases}
    &\hspace{5pt}
    v^{\mathit{fun}}(t) &\eqdef
    \begin{cases}
        v(t) &\text{if } v(t) \in \Z \\
        0^\mathit{fun}    &\text{otherwise}
    \end{cases}
\end{align*}
The denotation of~$\Susf$ requires all its terms to be defined,
following the intuition of strict sum described above.
In contrast, the denotation of~$\Sumf$ uses the neutral element~$0^\sumf = 0$ to evaluate undefined terms.
This has the same effect as discarding undefined terms from the sum, as done in ASP.
We can now explain the difference between~$\Susf$ and~$\Sumf$ in formal terms.
Let us reconsider constraint atoms~\eqref{eq:flingo-sumf-example} and~\eqref{eq:flingo-susf-example}.
On the one hand, we have~$v^{\sumf}(x_1) = v(x_1) = 1$ but~$v^{\sumf}(x_2) = 0$ because~$v(x_2$) is undefined.
Hence, for atom~\eqref{eq:flingo-sumf-example}, we have~$1 + 0 \leq 3$, and thus~$v \in \den{\sumf\{x_1 ; x_2\} \,\Leq\, 3}$.
In contrast,
for atom~\eqref{eq:flingo-susf-example},
we get $v(x_2) \notin \Z$ and, thus, we immediately have~$v \notin \den{\susf\{x_1 ; x_2\} \,\Leq\, 3}$.
As mentioned above, $\minf$ always behaves in a non\nobreakdash-strict way,
and accordingly, their denotations use their neutral elements to evaluate undefined terms.
$\Deff(x)$ expresses that the variable~$x$ is defined.

We now define the translation of \flingo\ programs into sets of formulas in \HTC.
Recall that \flingo\ constraint atoms have conditional terms of the form~$s : l_1, \ldots, l_n$,
which do not have the syntactic form of~\HTC\ conditional terms.
Hence, the translation of \flingo\ programs into \HTC\ needs to replace \flingo\ conditional terms with \HTC\ conditional terms.
This translation is context-sensitive, as it is determined by the specific constraint atom containing the conditional term.
Hence, for each \flingo\ term~$t$,
we define a corresponding term~$[t]^\mathit{fun}$ in \HTC\
where~$\mathit{fun}$ is one of the functions~$\sumf$, $\susf$, $\maxf$, or $\minf$.
This translation is defined as follows:
\begin{align*}
    [t]^\mathit{fun} \ &\eqdef \ t &&\text{if } t \text{ is a product term }
    \\
    [t]^\mathit{fun} \ &\eqdef \ \cterm{s\,}{\,0^\mathit{fun}\,}{\,\tau l_1 \land \ldots \land \tau l_n} &&\text{if } t \text{ is of the form } s : l_1, \ldots, l_n
\end{align*}
%
Whenever the condition is false, each term is evaluated to the corresponding neutral element of the function.
%
%
Next, we define the translation~$\mu$ of \flingo\ atoms into \HTC\ atoms.
For each atom~$c$ that is propositional or of the form~$\Deff\{x\}$, its translation~$\mu c$ is just~$c$ itself.
For any \flingo\ constraint atom~$c$ of the form
\texttt{\&}$\mathit{fun}\{t_1 ; \ldots ; t_n\} \prec s$
%
%
with~$\mathit{fun}$ among~$\{\sumf,\, \susf,\, \maxf,\, \minf\}$, we define~$\mu c$ as
\begin{gather}
    \mathit{fun}\{ \; [t_1]^\mathit{fun} ; \ldots ;  [t_n]^\mathit{fun} \; \} \prec s \ .
    \label{eq:tau.flingo-atom}
\end{gather}
The following proposition shows that~$\sumf$ and~$\susf$ only differ when they contain undefined integer variables, that is,
they are equivalent when all their integer variables are defined.
%
\begin{proposition}\label{prop:sumf-susf}
  Let~$c_1$ and~$c_2$ stand for the \flingo\ constraint atoms
  \begin{align*}
    \Sumf\{t_1 ; \ldots ; t_n\} \prec s
    \qquad\qquad
    \Susf\{t_1 ; \ldots ; t_n\} \prec s
  \end{align*}
  and let ~$\tuple{h,t}$ be an ht\nobreakdash-interpretation satisfying~$\inth(x)$
  for each integer variable~$x$ in~$\vars{c_1}$ and~$\vars{c_2}$.
  Then, $\tuple{h,t} \models \mu c_1$ if and only if $\tuple{h,t} \models \mu c_2$.
\end{proposition}


We extend the translation~$\mu$ to \flingo\ literals in the usual way, that is, $\mu\, \Not\ c$ is~$\neg\mu c$, and~$\mu\, \Not\ \Not\ c$ is~$\neg\neg\mu c$.
Finally, we extend the translation~$\mu$ to \flingo\ rules~\eqref{eq:flingo-rule} as
\begin{gather}
    \mu l_1 \land \ldots \land \mu l_n \to \mu a
    \label{eq:tau.flingo-rule}
\end{gather}
For a set~$P$ of \flingo\ rules, we define~$\fl(P)$ as the set of \HTC\ formulas containing formula~$\mu r$
for each rule~$r \in P$ as well as formulas of the form
\begin{align}
    \Deff(p) &\to \p &&\quad \text{for each propositional variable } p \in \X^p
    \tag{\ref{eq:clingcon-defh}}
    \\
    \Deff(x) &\to \inth(x)    &&\quad \text{for each integer variable } x \in \X^n
    \label{eq:flingo-inth}
\end{align}
%
Formulas~\eqref{eq:clingcon-defh} and~\eqref{eq:flingo-inth} ensure that variables only take their intended values:
propositional variables are either undefined or take the value~$\true$,
while integer variables must take an integer value when defined.
The formulas for propositional variables in~\eqref{eq:clingcon-defh} are the same as in \clingcon,
while the ones for integer variables in~\eqref{eq:flingo-inth} are weaker than in~\eqref{eq:clingcon-inth},
which ensures that integer variables are assigned integer values when defined,
but crucially they remain undefined otherwise.
Recall that in \clingcon, the formulas in~\eqref{eq:clingcon-inth} ensure that integer variables are always defined.

Syntactically, every \clingcon\ program~$P$ is a \flingo\ program and, thus,
we can also interpret it as a \flingo\ program by using the corresponding translation~$\fl(P)$.
Furthermore, every \clingcon\ atom~$c$ has no conditional terms, and, thus, $\mu c$ is just~$c$ itself.
Hence, for a \clingcon\ program~$P$, there are two differences between~$\fl(P)$ and~$\cl(P)$:
the denotation of~$\Sumf$ atoms and
that the latter uses the formulas in~\eqref{eq:clingcon-defh} for integer variables,
while the former uses the ones in~\eqref{eq:flingo-inth}.

While the denotation of~$\Sumf$ atoms in \clingcon\ coincides with that of~$\Susf$ atoms in \flingo,
we maintain a deliberate syntactic distinction.
This choice is driven by our goal to align \flingo\ as closely as possible with standard ASP aggregate conventions.
In particular, since standard ASP aggregates adhere to non-strict semantics,
\flingo\ adopts a notation that reflects this established behavior.
For instance, the \flingo\ program
\begin{lstlisting}[language=clingo,numbers=none,mathescape]
a :- &sum{ x : p } = 0.            p.
\end{lstlisting}
has a single stable model in which~$\code{a}$ and~$\code{p}$ are true and~$\code{x}$ is undefined.
However, if we replace~$\Sumf$ with~$\Susf$, the resulting program leaves~$\code{a}$ undefined.
This program can be rewritten in ASP using the aggregate~$\code{\#sum}$ as follows:
\begin{lstlisting}[language=clingo,numbers=none,mathescape]
a :- #sum{ X : val(X), p } = 0.    p.
\end{lstlisting}
The unique stable model of this program is~$\{ \code{a},\code{p} \}$ (as all instances of \code{val(X)} are false).

To illustrate the use of strict versus non-strict sums for knowledge representation, we revisit next  our running example about tariffs from the introduction.
A possible representation of the example in \flingo\ could be:
\begin{lstlisting}[language=clingo,numbers=left,mathescape]
&sum{tariff(cars,ca)}=25.
&sum{tariff(P,eu)}=15 :- sales(P,eu,_),
                         not &sus{tariff(P,eu)}!=15.
&sum{tariff(steel,eu)}=0.
&sum{tariff(aircraft,eu)}=25.
&sum{Y*tariff(P,C),P,C:sales(P,C,X),Y=X/100}=:taxincome.
\end{lstlisting}
\noindent Notice how the rule for expressing the default value (lines 2-3) uses $\Susf$ in the body to check that \code{tariff(P,eu)} does not have \emph{some specific value} different from 15.
If we used $\Sumf$ instead, an undefined tariff would count as $0^\sumf=0$ in the sum, and this is always different from 15, so the rule would not be applied, leaving the tariff undefined.
Notice, however, that in the computation of the \code{taxincome} (line 6), it is more convenient to use $\Sumf$ rather than $\Susf$.
This is because we might have facts like \code{sales(food,ca,10000)} without having defined a tariff for Canadian food in the database.
By using $\Sumf$, an undefined tariff in the aggregate just counts as 0 in the total sum (as would also happen in our standard ASP representation of the problem without constraints).

The following proposition shows the relation between the ht-models and stable models of~$\cl(P)$ and~$\fl(P^*)$
where $P^*$ is the result of replacing each occurrence of~$\Sumf$ with~$\Susf$ in a \clingcon\ program~$P$.
%
\begin{proposition}\label{prop:clingcon.ht.models.flingo}
  Let~$P$ be a \clingcon\ program and~$P^*$ be the result of replacing each occurrence of~$\Sumf$ with~$\Susf$ in~$P$.
  Then,
  \begin{enumerate}[label=(\alph*)]
  \item\label{item:ht:prop:clingcon.ht.models.flingo}
    every ht\nobreakdash-model of~$\cl(P)$ is also an ht\nobreakdash-model of~$\fl(P^*)$; and

  \item\label{item:stable:prop:clingcon.ht.models.flingo}
    every stable model of~$\fl(P^*)$ without undefined integer variables is also a stable model of~$\cl(P)$.
  \end{enumerate}
\end{proposition}

The relationship between \flingo\ and \clingcon\ described in
Proposition~\ref{prop:clingcon.ht.models.flingo}\ref{item:stable:prop:clingcon.ht.models.flingo}
parallels the one between stable and classical models,
with the caveat that we must restrict our focus to stable models devoid of undefined integer variables.
Just as every stable model is a classical model,
though not vice versa,
the presence of undefined variables in \flingo\ allows for stable models that lack a counterpart in \clingcon.
To illustrate,
consider the \clingcon\ program $\{\code{a} \ \code{:-} \ \Sumf\{x \} = x\}$ versus
the corresponding \flingo\ program $\{\code{a} \ \code{:-} \ \Susf\{x \} = x\}$.
The former yields infinitely many stable models where~$\code{a}$ is true and~$x$ is assigned an arbitrary integer.
In contrast, the \flingo\ program results in a unique stable model where~$\code{a}$ is false and~$x$ remains undefined.

Continuing with the analogy between stable and classical models,
in the latter, we can capture all models of a propositional theory by adding choice rules for each propositional variable.
Similarly, we can capture all stable models of a \clingcon\ program by
adding facts of the form~${\Sumf\{x\} = x}$ (aka $x=x$) for each integer variable~$x$.
These facts act as a kind of choice rule for integer variables, allowing them to pick any value.
%
\begin{proposition}\label{prop:clingcon.as.flingo}
  Let~$P$ be a \clingcon\ program and
  let~$F$ be the set of facts consisting of~${\Sumf\{x\} = x}$ for each integer variable~$x \in \X^n$.

  Then, the set of ht- and stable models of~$\cl(P)$, $\fl(P^*\cup F)$, and~$\fl(P \cup F)$ coincide.
\end{proposition}

We can further push this analogy by showing that \flingo\ allows a form of non\nobreakdash-monotonic reasoning for integer variables that is impossible in \clingcon.
Note that \clingcon\ is not monotonic in the sense of propositional logic because it is an extension of the stable model semantics.
Thus, we can just write a program containing propositional rules that is non\nobreakdash-monotonic.
However, we can show its non\nobreakdash-monotonicity comes only from the propositional part of the program, while the integer part is monotonic.
%
\begin{proposition}\label{prop:clingcon-monotonicity}
  Let~$P$ and~$F$ be two \clingcon\ programs such that~$F$ has no propositional variables.
  Then, every stable model of~$\cl(P\cup F)$ is also a stable model of~$\cl(P)$.
\end{proposition}

The language of \flingo\ is not monotonic in the sense of Proposition~\ref{prop:clingcon-monotonicity}.
For instance, consider the program consisting of the single rule~${\code{x} = 1 \ \code{:-} \ \Susf\{ \code{y} \} = 1}$, whose unique stable model leaves both~$x$ and~$y$ undefined.
If we add the fact~${\Susf\{ \code{y} \} = 1}$, which does not contain propositional variables,
then the resulting program's unique stable model assigns~$1$ to both~$x$ and~$y$.

%

%
In addition to the basic constraint atoms defined above, \flingo\ also supports~$\Maxf$ constraint atoms of the form~$\Maxf\{t_1 ; \ldots ; t_n\} \prec s$,
which are syntactic sugar for~$\Minf\{-t_1 ; \ldots ; -t_n\} \succ -s$ where~$\succ$ is the dual operator of~$\prec$
defined as follows:
\begin{align*}
    \succ \ &\text{ is } \
    \begin{cases}
        \ \Geq  &\text{if } \prec \text{ is } \Leq \\[-1ex]
        \ \Leq  &\text{if } \prec \text{ is } \Geq \\[-1ex]
        \ \Grt  &\text{if } \prec \text{ is } \Low \\[-1ex]
        \ \Low  &\text{if } \prec \text{ is } \Grt \\[-1ex]
        \ \prec &\text{otherwise }
    \end{cases}
\end{align*}
A common pattern in \flingo\ programs is to allow integer variables to take values within a given range.
This can be expressed using \emph{choice rules} for integer variables of the form
\begin{gather}
    \code{\&in}\{ s_1\code{..} s_2 \}  \ \Ass \ x \ \ruleo \ \ l_1, \ldots, l_m
    \label{eq:flingo-choice}
\end{gather}
where~$s_1$ and~$s_2$ are product terms, $x$ is an integer variable and~$l_1, \ldots, l_m$ are \flingo\ literals.
Formally,
a choice rule as in~\eqref{eq:flingo-choice} is an abbreviation for the rules
\begin{gather}
    \Susf\{ s_1 \}  \ \code{<=} \ x \ \ruleo \ \ l_1, \ldots, l_m, \Deff(x_1), \dotsc, \Deff(x_k)
    \label{eq:flingo-choice.lower}
    \\
    \Susf\{ s_2 \}  \ \code{>=} \ x \ \ruleo \ \ l_1, \ldots, l_m, \Deff(x_1), \dotsc, \Deff(x_k)
    \label{eq:flingo-choice.upper}
\end{gather}
where~$x_1, \ldots, x_k$ are all the integer variables occurring in~$s_1$ and~$s_2$.

Besides this syntactic sugar, \flingo\ also supports another common pattern for defining and constraining integer variables that work on the rule level.
This pattern consists on defining a variable in terms of other variables.
Constraint atoms~(\ref{eq:flingo-sumf}--\ref{eq:flingo-deff}) do not make any difference between the variables occurring on them.
Assignment rules allow us to naturally express this pattern.
An \emph{assignment rule} is an expression of the form%
\footnote{Note that the assigned variables are the ones occurring on the right hand side of the assignment operator.
  This is due to the current syntax of \clingo\ theory atoms,
  where operators can only appear on the right-hand-side of a constraint atom.}
\begin{gather}
    \code{\&}\mathit{fun}\{t_1 ; \ldots ; t_n\} \ \Ass \ s \ \ \ruleo \ \ \l_1, \ldots, l_m
    \label{eq:flingo-assignment}
\end{gather}
where each~$t_i$ is a \flingo\ term, $s$ is a product term,
$\mathit{fun}$ is one of the operations~$\sumf$, $\susf$, $\maxf$, or $\minf$,
and each~$l_i$ is a \flingo\ literal.
Such rules only define the variables in~$s$,
so that the constraint atom obtained by replacing the assignment operator~$\Ass$ with equality~$\Equ$ is satisfied.
However, it does not define any variables in~$t_1, \ldots, t_n$.
For instance, the assignment rule~$\code{\Susf\{$x$ ; $y$\} \Ass\ $z$}$ ensures that~$z$ is defined as the sum of~$x$ and~$y$,
whenever both~$x$ and~$y$ are defined.
However, if~$x$ or~$y$ are undefined, then the rule has no effect on the value assigned to~$z$.
So, if we add facts~${\code{$x$ = 1}}$ and~${\code{$y$ = 2}}$ to the above assignment,
then the resulting program has a single stable model where~$x$, $y$, and~$z$ are assigned~$1$, $2$, and~$3$, respectively.
If instead,
we add only the fact~${\code{$x$ = 1}}$,
then the resulting program has a single stable model where~$x$ is assigned~$1$, while~$y$ and~$z$ are undefined.
And if we add instead the facts~${\code{$x$ = 1}}$ and~${\code{$z$ = 5}}$,
then the resulting program has a single stable model where~$x$ and~$z$ are assigned~$1$ and~$5$, respectively,
while~$y$ is undefined.

Formally,
an assignment rule as in~\eqref{eq:flingo-assignment} where each~$t_i$ is a product term is an abbreviation for the rule:
\begin{gather}
    \code{\&}\mathit{fun}\{t_1 ; \ldots ; t_n\} \ \Equ \ s \ \ \ruleo \ \ \ l_1,\, \ldots, l_m,\, \Deff(x_1), \dotsc, \Deff(x_k)
    \label{eq:flingo-assignment-abbr}
\end{gather}
where~$x_1, \ldots, x_k$ are all the integer variables occurring in~$t_1,\dotsc, t_n$.
When some~$t_i$ is a conditional term, conditions are first removed as explained in the next section.

%

\section{The \flingo\ system}
\label{sec:flingo:system}

We have implemented a new system called \flingo\ as an extension of \clingo~\citep{gekakaosscwa16a}
with the integer constraint atoms as described in Section~\ref{sec:approach}.
The syntax uses \clingo\ theory atoms starting with~`\texttt{\&}' to denote integer constraint atoms, and
grounding of logic program variables is handled using \clingo's native capabilities.
Logic program variables are different from integer variables.
Such variables start with a lowercase letter and can only occur inside integer constraint atoms.
As usual in ASP, logic program variables start with an uppercase letter and can be used anywhere in the program.
Logic program variables are grounded before the solving process starts,
while integer constraint variables are handled during solving.
\Flingo\ uses \clingcon\ as backend for handling integer constraints during solving.
A grounded \flingo\ program is translated into a \clingcon\ program by a series of rewriting transformations
detailed below.
These transformations are based on ideas
to compile away aggregates with constraints~\citep{cafascwa20a} and
to compile integer constraints allowing undefined integer variables into standard integer constraints~\citep{cakaossc16a},
but adapted to the richer setting of~\flingo.

Our transformation consists of eight steps.
The \emph{first step} is to replace
in the scope of a constraint atom whose operation is not~$\Susf$,
each conditional term  of the form~${s : l_1, \ldots, l_m}$ by
\[
  s : l_1, \ldots, l_m,\, \Deff(x_1), \dotsc, \Deff(x_k)
\]
where~$x_1, \ldots, x_k$ are all the integer variables occurring in~$s$.
Similarly, a product term~$s$ is replaced by the conditional term~$s : \Deff(x_1), \dotsc, \Deff(x_k)$.
After that, all occurrences of~$\Sumf$ atoms are replaced by~$\Susf$ atoms.

The \emph{second step} consists of replacing all conditional terms of the form~\eqref{eq:conditional.term}
by conditional terms of the form~${s : a}$ where~$a$ is a propositional atom.
This is done by introducing fresh propositional atoms~$a$ to represent the condition and
adding rule~$a \  \code{:-} \ l_1, \ldots, l_m$ to define this fresh atom.

%
The \emph{third step} consists of removing conditions in constraint atoms.
We replace each conditional constraint term of the form~${s : a}$ by a fresh integer variable~$y$, and we add rules
\begin{align*}
  \Susf\{s\} = y \ &\code{:-} \ a,\, \Deff(x_{1}), \dotsc, \Deff(x_{k})
  \\
  \Susf\{s\} = y \ &\code{:-} \ a,\, \Deff(y)
  \\
  \Susf\{ 0^{\mathit{fun}}\} = y   \ &\code{:-} \ \code{not } a
  \\
  \{a\}            \ &\code{:-} \ \Deff(y)
\end{align*}
where~$x_{1},\dotsc,x_{k}$ are all integer variables in~$s$, and~$0^{\mathit{fun}}$ depends on the operation~$\mathit{fun}$ where the conditional constraint term occurs as described in Section~\ref{sec:approach}.

The \emph{fourth step} is to rewrite all abbreviations described in the previous section: assignment and choice rules, and all occurrences of~$\texttt{max}$ constraint atoms.

The \emph{fifth step} is to replace each constraint atom of the form~${\texttt{\&}\mathit{fun}\{s_1  ; \ldots ; s_n \} \prec s}$
by a fresh propositional atom~${\texttt{\&}\mathit{fun}(\mathit{hb})\{s_1  ; \ldots ; s_n \} \prec s}$
where~$\mathit{hb}$ is either~$\code{head}$ or~$\code{body}$
depending on whether the atom occurs in the head or the body, respectively.
\Clingcon\ only recognizes constraint atoms whose operation is~$\Sumf$, thus these atoms with operation~${\texttt{\&}\mathit{fun}(hb)}$ are treated as regular propositional atoms by \clingcon.
We add later rules that relate the value of the fresh propositional atoms and the integer variables.

The \emph{sixth step} is to define $\Minf$ atoms in terms of $\Susf$ atoms as follows.
For every constraint atom of the form~$\Minf(hb)\{s_1 ; \ldots ; s_n\} \prec s$,
we add the rule
\begin{align*}
  &\code{:-} \ \mathit{def}, \ \code{not} \ \mathit{member}
  \\
  &\Susf(\code{head})\{ \smax_\cl \} = m \ \code{:-} \ \code{not} \ \mathit{def}
\end{align*}
where~$m$ is a fresh integer variable representing the minimum of~$s_1,\dotsc,s_n$,
and~$\mathit{def}$ and~$\mathit{member}$ are fresh propositional atoms defined by the following rules for each~${i=1,\dotsc,n}$:
\begin{align*}
  &\mathit{def} \ \code{:-} \ \Deff(x_{i_1}), \dotsc, \Deff(x_{i_k})
  \\
  &\Deff\{m\} \ \code{:-} \ \mathit{def}
  \\
  &\code{:-} \ \Susf(\code{body})\{ s_i \} \ \Low \ m
  \\
  &\mathit{member} \ \code{:-} \ \Susf(\code{body})\{ s_i \} \ \Equ \ m
\end{align*}
with~$x_{i_1},\dotsc,x_{i_k}$ are all integer variables occurring in~$s_i$.
In addition, if~$hb$ is~$\code{head}$, we add the rule
\begin{align*}
  \Susf(\code{head})\{ m \} \prec s
  \ &\code{:-} \
      \Minf(\code{head})\{s_1 ; \ldots ; s_n\} \prec s
\end{align*}
while if~$hb$ is~$\code{body}$, we add the rule
\begin{align*}
  &\Minf(\code{body})\{s_1 ; \ldots ; s_n\} \prec s \ \code{:-} \ \Susf(\code{body})\{ m \} \prec s
\end{align*}

In the \emph{seventh step}, we relate the fresh propositional atoms introduced in step five with the integer variables.
For every integer variable~$x$ occurring in the program, we add
\begin{align*}
  \Sumf\{0\} = x \ \code{:-} \ \code{not } \Deff(x)
\end{align*}
This rule ensures that undefined integer variables are assigned a unique value and, thus, avoids multiple stable models that only differ in the value assigned to undefined integer variables.
For every constraint atom~$\Susf(\code{head})\{s_1 ; \ldots ; s_n\} \prec s$, we add the rules
\begin{align*}
  &\code{:-} \ \Susf(\code{head})\{s_1 ; \ldots ; s_n\} \prec s, \ \code{not } \Sumf\{s_1 ; \ldots ; s_n\} \prec s
  \\
  &\mathit{aux} \ \code{:-} \ \Susf(\code{head})\{s_1 ; \ldots ; s_n\} \prec s
  \\
  &\Deff(x_1) \ \code{:-} \ \mathit{aux}
  \qquad\qquad\dots\qquad\qquad
  \Deff(x_k) \ \code{:-} \ \mathit{aux}
\end{align*}
where~$x_1,\dotsc,x_k$ are all integer variables occurring in~$s_1,\dotsc,s_n, s$, and~$\emph{aux}$ is a fresh propositional atom.
For every constraint atom of the form~$\Susf(\code{body})\{s_1 ; \ldots ; s_n\} \prec s$, we add the rule
\begin{align*}
  &\Susf(\code{body})\{s_1 ; \ldots ; s_n\} \prec s \  \code{:-} \ \Sumf\{s_1 ; \ldots ; s_n\} \prec s, \Deff(x_1), \dotsc, \Deff(x_k)
\end{align*}
where~$x_1,\dotsc,x_k$ are all integer variables occurring in~$s_1,\dotsc,s_n,s$.

The \emph{final step} replaces each atom of the form~$\Deff(x)$ by a new propositional atom~$\texttt{def(x)}$,
so \clingcon\ does not assume these atoms to be in the scope of a choice rule.

Every step of this transformation is feasible in linear time with respect to the size of the program.
Thus, the overall transformation is linear as well.
%

%
\Flingo\ supports the same useful modeling features for the ASP methodology as \clingo, such as choice rules, defaults, and aggregates, now lifted to linear integer arithmetic.
In Listing~\ref{lst:flingo-example}, we show a \flingo\ program that models the configuration of a bike that consists of a frame and an optional bag, and the calculation of the total price.
%
\begin{lstlisting}[language=clingos, caption={Bike configuration with \flingo.}, label={lst:flingo-example}, basicstyle=\footnotesize\ttfamily]
price(frame,15).                  default_range(1,2).
select(frame).                  { select(bag) }.
&sus{V} = price(P)    :- select(P), price(P,V).
&in{L..U} =: price(P) :- select(P), default_range(L,U)
                         not &sus{price(P)}<L, not &sus{price(P)}>U.
&sus{price(P) : select(P)} =: price(total).
\end{lstlisting}
%
Lines~1 and~2 define the problem instance.
%
Lines~3-6 provide the general problem encoding.
Line~3 defines the price of a selected component as the one in the instance, when provided.
Lines~4-5 define the price of a selected component using the default range defined in the instance:
unless the price is provably outside the default range, the price of a selected component is defined by the default range.
%
%
Finally, Line~6 defines the total price as the sum of the prices of the selected components.
%
%
%
This example illustrates how \flingo\ allows for a natural encoding of configuration problems with numeric attributes, where defaults, choices, and aggregates are common modeling features.
\Flingo\ has been used to model and solve configuration problems formulated in the \textsc{Coom} configuration language~\citep{baheosreruscwa26a},
where the empirical evaluation demonstrates its effectiveness in solving configuration problems with numeric attributes featuring large domains, where it significantly outperforms~\clingo.
%

%


\section{Discussion}\label{sec:discussion} 

We have introduced \flingo,
a novel CASP system designed to bridge the gap between high-level ASP modeling and
the operational requirements of numeric constraint solving.

Our contributions can be summarized as follows:
(i)
We presented a logic programming language that integrates default values, choice rules, and aggregates directly into constraint atoms.
This preserves the standard ASP ``look and feel'' while leveraging the power of numeric backends.
(ii)
We defined the semantics of \flingo,
distinguishing between strict (\lstinline{&sus}) and non-strict (\lstinline{&sum}) operations to accommodate both traditional CASP behavior and ASP-style aggregates.
(iii)
We developed a translation from \flingo\ to \clingcon,
combining the effectiveness of state-of-the-art CASP solvers with ASP declarativeness.

\cite{lierler23a} analyzes an extensive list of CASP systems from the literature, including the already mentioned \clingcon.
None of these systems described there allows for leaving constraint variables undefined, representing default values, or the use of aggregates involving both constraints variables and arbitrary conditions.
Perhaps the closest system to \flingo\ regarding these features is \clingof\footnote{\url{https://mbal.asklab.net/clingof}}~\citep{balduccini13a}, which deals with non-Herbrand numerical functions, and allows for undefinedness and default values.
%
However, \clingof\
does not perform numerical constraint solving.
System s(CASP)~\citep{arcasamagu18a} also mixes ASP with numerical constraints, but unlike standard CASP approaches
based on grounding and solving, it follows a goal-directed evaluation as in Prolog.
Under this paradigm, problems are modeled and solved in a substantially different way, so transferring examples from the ASP/CASP literature to s(CASP) is not always straightforward.
%
%
If we wanted to extrapolate s(CASP) constraints to grounding-based CASP, their use would actually occur during the grounding phase, rather than in the solving phase as happens in the rest of CASP systems.
Finally, aggregates are not directly representable inside s(CASP) constraints, but must be encoded through other Prolog predicates.

While the current implementation of \flingo\ relies on \clingo's theory grammar,
future work will focus on making the input language more natural and
exploring further optimizations in the translation process to enhance scalability.
The \flingo\ system is publicly available at \url{https://github.com/potassco/flingo}.


%
\paragraph{Acknowledgments}
We would like to thank anonymous reviewers for their valuable feedback that allowed us to improve presentation of several points.
This work was supported
by grant PID2023-148531NB-I00 funded by Spanish
Ministry MCIU/AEI/10.13039/501100011033, funds FEDER, EU,
by the National Science Foundation CAREER award 2338635, USA,
and by DFG grant SCHA 550/15, Germany.
Any opinions, findings, and conclusions or recommendations expressed in this material are those of the authors and do not necessarily reflect the views of the National Science Foundation.
\bibliographystyle{plainnat} 
\bibliography{include/bibliography/krr,./local,include/bibliography/procs}

\end{document}